\documentclass[sigconf, noend]{acmart}
\AtBeginDocument{%
  \providecommand\BibTeX{{%
    \normalfont B\kern-0.5em{\scshape i\kern-0.25em b}\kern-0.8em\TeX}}}

\setcopyright{acmcopyright}
\copyrightyear{2023}
\acmYear{2023}
\acmDOI{XXXXXXX.XXXXXXX}

\acmConference[MM '23]{the 31th ACM International Conference on Multimedia}{October 28--November 03,
  2023}{Ottawa, Canada}
\acmPrice{15.00}
\acmISBN{978-1-4503-XXXX-X/18/06}

\usepackage{multirow}
\usepackage{graphicx}
\usepackage{subfigure}
\usepackage{amsmath}
\PassOptionsToPackage{prologue,dvipsnames}{xcolor}
\usepackage{bbding}
\usepackage{pythonhighlight}%插入python代码
\usepackage[ruled]{algorithm2e}

\usepackage{listings}
\usepackage{xcolor}
\definecolor{mygreen}{rgb}{0,0.6,0}
\definecolor{mygray}{rgb}{0.5,0.5,0.5}
\definecolor{mymauve}{rgb}{0.58,0,0.82}

\begin{document}

\title{DRPT: Disentangled and Recurrent Prompt Tuning for Compositional Zero-Shot Learning}


\author{Xiaocheng Lu}
\authornote{Both authors contributed equally to this research.}
\email{xiaochenglu1997@gmail.com}
\author{Ziming Liu}
\authornotemark[1]
\email{ziming.liu@connect.polyu.hk}
\affiliation{%
  \institution{Department of Computing \\ Hong Kong Polytechnic University}
  % \streetaddress{P.O. Box 1212}
  \city{Hong Kong SAR}
  \country{China}
}

\author{Song Guo}
\email{song.guo@polyu.edu.hk}
\affiliation{%
  \institution{Department of Computing \\ Hong Kong Polytechnic University}
  % \streetaddress{1 Th{\o}rv{\"a}ld Circle}
  \city{Hong Kong SAR}
  \country{China}}

\author{Jingcai Guo}
\email{jingcai.guo@connect.polyu.hk}
\affiliation{%
  \institution{Department of Computing \\ Hong Kong Polytechnic University}
  \city{Hong Kong SAR}
  \country{China}
}

\author{Fushuo Huo}
\email{fushuo.huo@connect.polyu.hk}
\affiliation{%
  \institution{Department of Computing \\ Hong Kong Polytechnic University}
  \city{Hong Kong SAR}
  \country{China}
}

\author{Sikai Bai}
\email{whitesk1973@gmail.com}
\affiliation{%
  \institution{Department of Computing \\ Hong Kong Polytechnic University}
  \city{Hong Kong SAR}
  \country{China}
}

\author{Tao Han}
\email{hantao10200@gmail.com}
\affiliation{%
  \institution{Shanghai AI Laboratory}
  \city{Shanghai}
  \country{China}
}
% \author{Anonymous Authors}
% \affiliation{
%   \institution{Paper ID: 2803}
% }

%\renewcommand{\shortauthors}{Xiaocheng Lu and Ziming Liu, et al.}
\renewcommand{\shortauthors}{Xiaocheng Lu and Ziming Liu et al.}

\begin{abstract}
Compositional Zero-shot Learning (CZSL) aims to recognize novel concepts composed of known knowledge without training samples. 
Standard CZSL either identifies visual primitives or enhances unseen composed entities, and as a result, entanglement between state and object primitives cannot be fully utilized. 
%How to make good use of the entanglement between state and object primitives is the bottleneck of standard CZSL.
%To achieve this capability, respectively identifying visual primitives or enhancing unseen composed entities are two mainstream solutions. 
%
Admittedly, vision-language models (VLMs) could naturally cope with CZSL through tuning prompts, while uneven entanglement leads prompts to be dragged into local optimum.
In this paper, we take a further step to introduce a novel \underline{D}isentangled and \underline{R}ecurrent \underline{P}rompt \underline{T}uning framework termed \textbf{\textit{DRPT}}\footnote{The code is attached to the supplementary material and will be released at https://github.com/Forest-art/DRPT-torch.git} to better tap the potential of VLMs in CZSL. 
Specifically, the state and object primitives are deemed as learnable tokens of vocabulary embedded in prompts and tuned on seen compositions. 
Instead of jointly tuning state and object, we devise a disentangled and recurrent tuning strategy to suppress the traction force caused by entanglement and gradually optimize the token parameters, leading to a better prompt space.
Notably, we develop a progressive fine-tuning procedure that allows for incremental updates to the prompts, optimizing the object first, then the state, and vice versa.
Meanwhile, the optimization of state and object is independent, thus clearer features can be learned to further alleviate the issue of entangling misleading optimization. 
Moreover, we quantify and analyze the entanglement in CZSL and supplement entanglement rebalancing optimization schemes.
\textbf{\textit{DRPT}} surpasses representative state-of-the-art methods on extensive benchmark datasets, demonstrating superiority in both accuracy and efficiency.
% high efficiency of our model.
% most recent state-of-the-art methods, demonstrating high efficiency of our model.

\end{abstract}






\maketitle

\section{Introduction}

How are humans able to recognize the vast number of novel concepts that have never been seen before?
This capability derives from the generalization of learned knowledge to unseen entities, even a non-existent concept such as \textit{"blue apple"} can be imagined in terms of \textit{"red apple"} and \textit{"blue bag"}.
To equip models with the same ability as humans do, a new category of zero-shot learning was recently proposed, namely Compositional Zero-Shot Learning (CZSL).
Specifically, CZSL aims to recognize unseen composed concepts (state + object, e.g. \textit{ripe apple}) based on seen primitive concepts during the training phase.
Furthermore, CZSL requires that the model be able to generalize well from known knowledge to novel compositions, as well as learn the internal relationship between state and object.

\begin{figure}[t]
  \centering
  \includegraphics[scale=0.69]{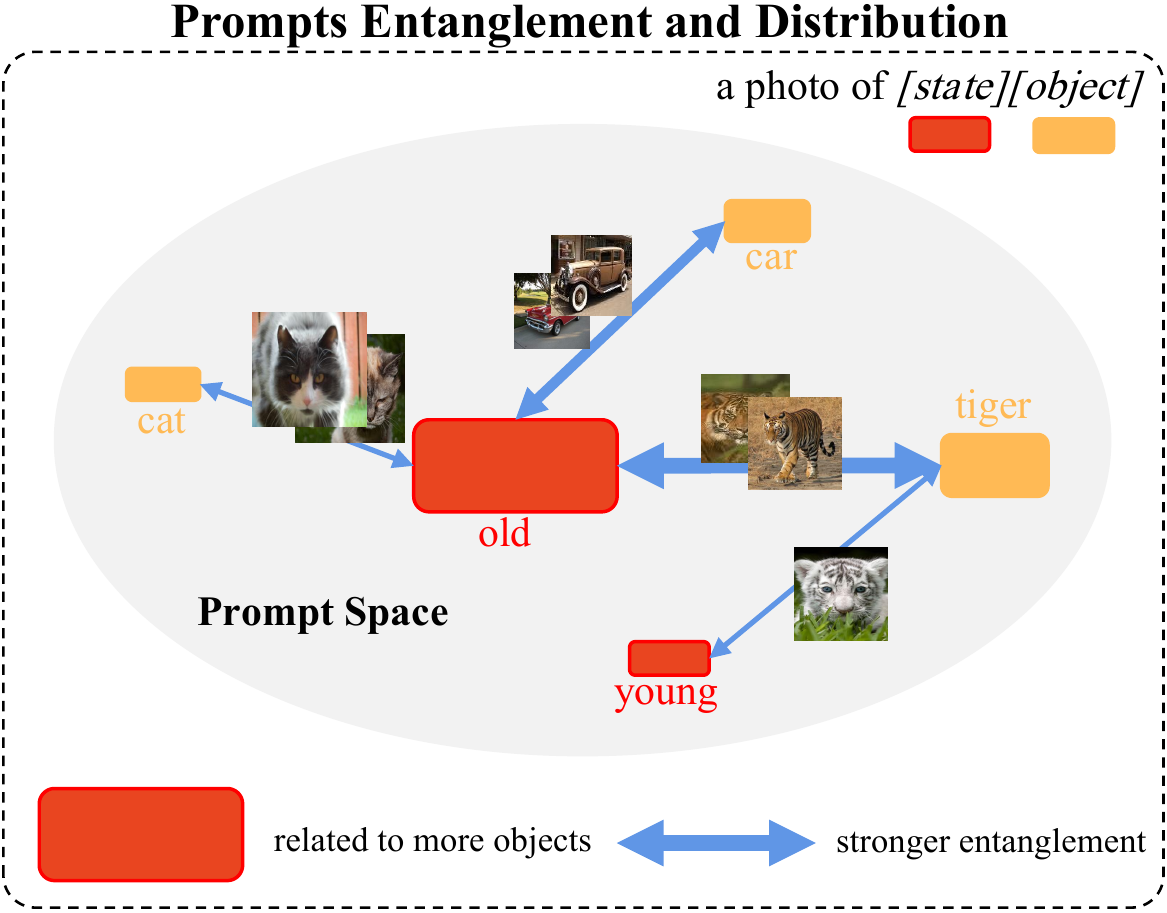}
   \caption{Prompts entanglement and distribution. 
   The intrinsic relationship between state and object tokens causes them to become entangled with each other when optimizing.
Moreover, the entanglement is not uniform. As shown in the figure, larger state prompt represents that it's related to more objects and vise versa.
Also, the thicker the line, the stronger the entanglement will be. Notably, prompt that is entangled with more prompts will have greater resistance to parameter updating, while keeping strong traction.}
   \label{fig:moti-update}
\end{figure}

\begin{figure*}[t]
  \centering
  \includegraphics[scale=0.65]{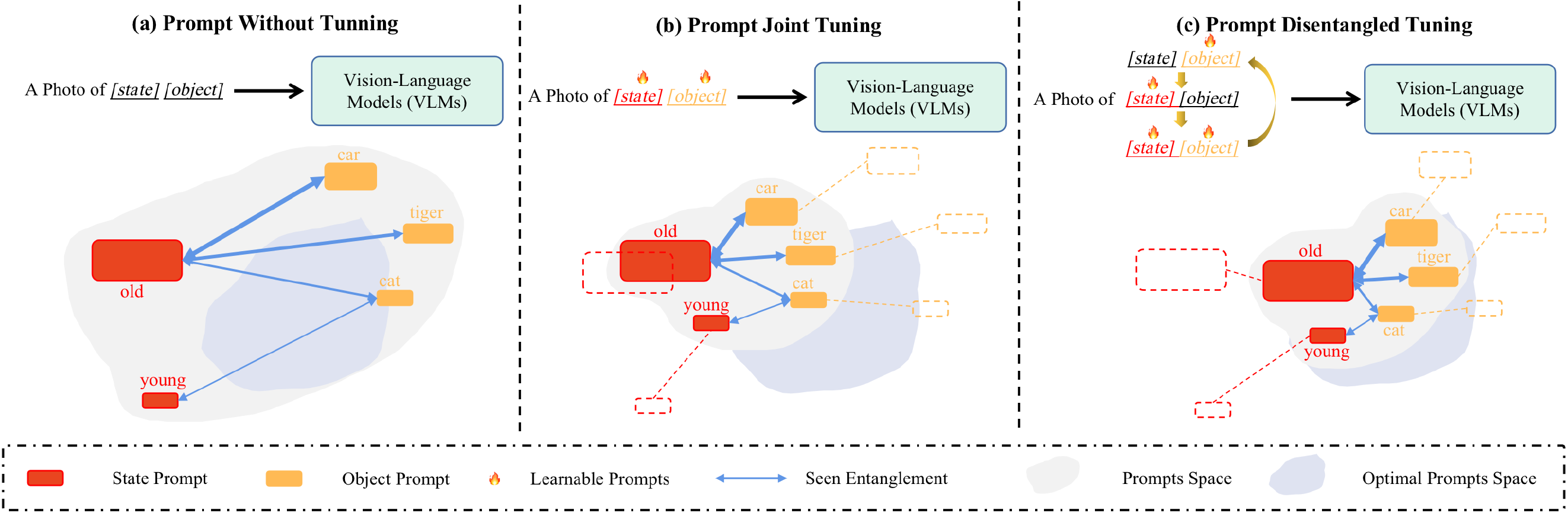}
   % \caption{Motivation behind our disentangled and recurrent tuning method. As the figure shown, state and object would tangle together to learn locally optimal parameters due to the strong intrinsic connection between them. Our method breaks down the learning process, independently learning the parameters of state and object, and then joint learning. When parameters are updated to a certain extent, the model will carry out this process periodically, making state and object jump out of the local optima.}
   \caption{Motivation and optimization diagram of three prompts tuning methods. As the figure shown, there are various entanglement between the prompts of state and object, causing them to gather in the area of high entanglement, that is, the learning of prompt joint tuning algorithm is inclined to this.
   And the existence of traction force of entanglement makes them deviate from the optimal parameter optimization and get stuck here.
   While large entangled prompts in prompt disentangled tuning method (\textbf{\textit{DRPT}}) resist traction and get updates to a greater extent, leading the prompts space to optimum.}
   \label{fig:moti-pull}
\end{figure*}

CZSL is characterized by entanglement between state and object, such as \textit{young} describing a \textit{tiger} instead of an \textit{apple}.
Existing methods~\cite{YongLuLi2020SymmetryAG, IshanMisra2017FromRW} construct dual separate classifiers to recognize state and object respectively, neglecting the intrinsic relation between them.
Additionally, graph neural network (GNN) is introduced in~\cite{MassimilianoMancini2021LearningGE} to learn the association between state and object.
To discard impossible compositions, an external knowledge network is utilized to compute the possibility score for overall compositions in~\cite{ShyamgopalKarthik2022KGSPKG}.
Some other approaches are to optimize the distance between visual features and composite embeddings in the semantic space~\cite{TusharNagarajan2018AttributesAO, ZhixiongNan2019RecognizingUA}.
All of these methods are based on visual features, and they either could not make proper use of entanglement between state and object, or fail to generalize well unseen concepts.

Recently, the first vision-language model (VLM) for CZSL has been proposed named CSP~\cite{nayak2022learning}, which introduces a composed prompt like \textit{"a photo of old cat"} and computes the similarities between images and prompts.
However, CSP makes the model easily fall into local optimum due to the joint training of primitive concepts and uneven entanglement distribution.
If we could utilize entanglement wisely, we believe that tokens of state and object could be co-optimized in prompts better.
As is shown in Fig.~\ref{fig:moti-update}, since there is entanglement between state and object and the entanglement is not evenly distributed, larger prompt which is related to more other prompts would have stronger traction force so that the strongly entangled prompts will get entangled and indirectly lead other prompts near them.
Conversely, large numbers of small entanglement prompts can also mislead the optimization of few large entanglement compositions.
In other words, uneven entanglement takes the model's parameter updates away from the optimal direction and allows these prompts to tangle together at a local optimum, as shown in Fig.~\ref{fig:moti-pull}(b).

Inspired by these algorithms, we propose a new novel VLMs based method, which is the first work to tackle CZSL from tuning, namely Disentangled and Recurrent Prompt Tuning (\textbf{\textbf{\textit{DRPT}} }) to improve VLMs for compositional zero-shot learning.
\textbf{\textit{DRPT}} treats states and objects as learnable composed concepts and embeds them in the general prompt like "\textit{a photo of} \textit{[state] [object]}".
In particular, we design a progressive tuning strategy to decouple the parameter updating for the prompts of state and object.
Specifically, previous joint optimization is shifted to periodic piecewise optimization in \textbf{\textit{DRPT}}, allowing state and object prompts to learn more independent and clean feature parameters during their respective update phases.
To overcome the barrier of entanglement, the state and object are separately optimized during training and then co-optimized to learn better parameters.
The whole optimization process is defined as three stages: \textit{object}, \textit{state}, and \textit{object+state}, which optimize the model with these three statuses successively in a chain optimization form.
Even if there is still a stage of state-object joint optimization, the initial parameters of state and object have already been disentangled and would be decomposed again in the next round.
This disentangled design allows them not to have strong entanglement in the separated tuning phase.
In contrast to the overuse of entanglement by CSP, \textbf{\textit{DRPT}}  prevents state and object from falling into local optimum due to the traction force of entanglement when parameters are updated as shown in Fig.~\ref{fig:moti-pull}(c).
Notably, \textbf{\textit{DRPT}}'s rotational fine-tuning design is simple, efficient, and enlightening.

The contributions of this paper can be summarized as follows: 1) \textbf{\textit{DRPT}} is the first to address CZSL from the perspective of prompt tuning and devise a recurrent fine-tuning approach for disentanglement, easing the obstacle to parameter updating between state and object due to traction force of entanglement to a great extent; 2) We are the first to quantify the entanglement of primitive concepts in CZSL, and study the impact of entanglement through experiments, showing that \textbf{\textit{DRPT}} can effectively alleviate entanglement problem; 3) Extensive experiments demonstrate that our progressive tuning strategy can significantly improve multiple metrics on three challenging datasets with various entanglement.

\section{Related Work}
Here, we describe related works about compositional zero-shot learning and 
prompt learning.

\textbf{Compositional Zero-Shot Learning (CZSL)}.
The point of CZSL~\cite{YananGu2021ClassIncrementalIS, YongLuLi2020SymmetryAG, TomasMikolov2013DistributedRO, IshanMisra2017FromRW, TusharNagarajan2018AttributesAO} is to learn the compositionality of state and object pairs based on seen compositions and generalize to the unseen composed concepts.
Unlike traditional Zero-Shot Learning (ZSL)~\cite{ JingcaiGuo2020ANP, ChristophHLampert2014AttributeBasedCF, XiangyuLi2021GeneralizedZL, YangLiu2021GoalOrientedGE, KunWei2020LifelongZL, guo2023graph, liu2023ml2p}, which uses multiple attributed vectors to identify unseen objects, concepts in CZSL exist as state-object pairs.

Prior algorithms could be classified into two mainstream directions.
Specifically, some methods learn a sole classifier to identify compositions and a transformation block to convert primitive concets~\cite{misra2017red}, which is inspired by~\cite{biederman1987recognition, hoffman1984parts}.
Yang \textit{et al.} try to learn disentangled and compositional concepts hierarchically~\cite{MuliYang2020LearningUC}.
Li \textit{et al.} establishes dual classifiers to recognize state and object respectively, also combines with contrastive learning to 
enforce separate recognition of state and object~\cite{li2022siamese, yang2023dual}.
Also, some methods complement the intrinsic relationship between state and object in the post-processing phase, such as graph neural network (GNN) in~\cite{MassimilianoMancini2021LearningGE} or external knowledge network in~\cite{ShyamgopalKarthik2022KGSPKG}.
In contrast to this class of methods, other algorithms treat the state-object pairing as an entity and learn the joint representation of state and object~\cite{YuvalAtzmon2020ACV, purushwalkam2019task, XinWang2019TaskAwareFG}.
They try to learn a modular module to reconstruct new unseen compositions based on seen compositions like <\textit{M(old cat, young tiger) -> old tiger}>~\cite{purushwalkam2019task, XinWang2019TaskAwareFG}.
Or they would learn a common embedding space where visual features and composed embeddings could optimize the distance between them~\cite{TusharNagarajan2018AttributesAO, ZhixiongNan2019RecognizingUA}.
Notably, these algorithms are all based on visual features, making it difficult to both dig the intrinsic relationship between primitive concepts and get a good sense of unseen compositions.

Unlike the above methods, Nihal \textit{et al.} propose CSP~\cite{nayak2022learning} based on vision language model (VLM), which accumulates the compositions of state and object into a vocabulary and replaces them into prompts. 
Also, Lu \textit{et al.} decompose the prompt features and fuse them with visual feature to enhance the sensitivity of unseen composition in the contrastive space~\cite{lu2022decomposed}.
Nevertheless, these methods overapply the entanglement between state and object, causing the parameters in prompts to fall into local optima during the optimization process.

\textbf{Prompt Learning.}
Prompt Learning aims to reformulate the input text by a specific template, and try to fully utilize the large-scale pre-trained language model on this task~\cite{bach2022promptsource, bommasani2021opportunities, brown2020language, sanh2021multitask, vu2021spot, zhou2022learning}. 
Unlike fine-tuning, which utilizes pretrained models for downstream tasks, prompt learning tries to adapt various downstream tasks to pre-trained models by reformulating them.
Prompting narrows the distance between the pre-trained model and downstream tasks, achieving great performance in zero-shot or few-shot on a wide range of tasks~\cite{GuanghuiQin2023LearningHT, AlecRadford2021LearningTV}.

Specifically, CLIP~\cite{AlecRadford2021LearningTV} is the original model of prompt learning, which has been pre-trained on nearly 400 million text-image pairs.
Using CLIP directly for classification tasks can achieve good results even in the case of training-free or zero-shot.
Based on this, Zhou \textit{et al.} propose CoOp~\cite{KaiyangZhou2021LearningTP}, transforming the prefix part of the prompt to a soft learnable context like "\textit{[v1][v2][v3]class}" and only fine-tune this prefix to adapt with downstream tasks.
CoCoOp~\cite{zhou2022conditional} introduces a Meta-Net into CoOp to learn dynamic prefix prompts, enhancing response for new classes.
\cite{li2022grounded, zhang2022glipv2} applied CLIP to the object detection task and deeply fuses text with image features.
These aforementioned algorithms, whether purely visual or based on visual language models (VLMs), all focus on the design of the model, ignoring the optimization process of parameters.

\begin{figure*}[t]
  \centering
  \includegraphics[scale=1.05]{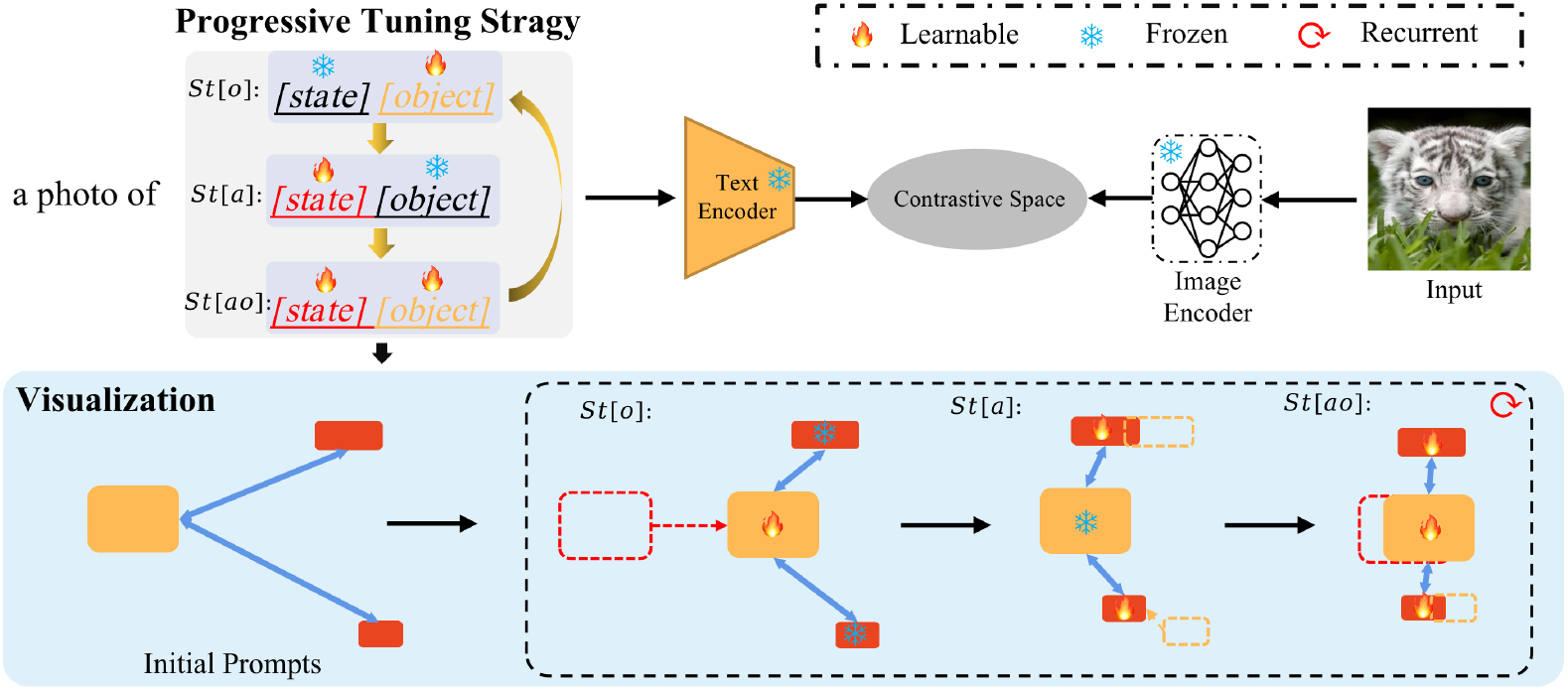}
   \caption{Overall framework of our method. Compositional prompts are converted as learnable tokens and tuned progressively with disentangled and recurrent tuning strategy. Specifically, \textbf{\textit{DRPT}} tune object vocabulary first, then state, and jointly tune them finally. Notably, the progressive tuning strategy could effectively suppress the traction force of entanglement and make different degrees of entanglement prompts can be effectively updated.}
   \label{fig:framework}
\end{figure*}

\section{Methodology}
In this section, we investigate traction force of entanglements and detail the learnable disentangled prompts, followed by recurrent tuning, namely \textbf{\textit{DRPT}}.
An overview of \textbf{\textit{DRPT}} is illustrated in Fig.~\ref{fig:framework}.

For compositional zero-shot learning, models aim to learn among seen compositions and identify the state and object of unseen concepts.
Previous mainstream algorithms constructed dual independent classifiers to identify state and object respectively, holistically neglecting the intrinsic relationship between them.
Instead, we focus on vision language models (VLMs), introducing both visual and textual modalities to tackle this task.
It is straightforward to establish combinations of state and object as prompts (e.g. \textit{a photo of old cat}) and then fine-tune them, like CSP~\cite{nayak2022learning}.
Nevertheless, due to the traction force of entanglement between a state and an object, learning the textual vector of a certain state may be affected by other objects and vice versa.
From the perspective of prompt tuning, we study and design a set of disentangled and recurrent tuning algorithm to suppress the powerful entanglement among textual prompts.

\subsection{Problem Formulation}
The goal of CZSL is to recognize compositions of two primitive concept sets, objects set $\mathcal{O} = \left \{ o_{0},o_{1},\dots ,o_{m}  \right \} $ and state/attribute set $\mathcal{A} = \left \{ a_{0},a_{1},\dots ,a_{n}  \right \} $ respectively.
And a set of size $n \times m$ would be denoted as $\mathcal{C} = \mathcal{A}  \times \mathcal{O}$, composing the state and object sets.
Besides, we define two disjoint sets $\mathcal{C}^{s}$ and $\mathcal{C}^{u}$, where $\mathcal{C}^{s}$, $\mathcal{C}^{u}$ are subsets of the composition set $\mathcal{C}$ and $\mathcal{C }^{s} \cap \mathcal{C}^{u} = \phi  $.
Specifically, $\mathcal{C}^{s}$, $\mathcal{C}^{u}$ represent the seen and unseen sets, respectively, where $\mathcal{C}^{s}$ is used for training and $\mathcal C^{u}\cup  \mathcal C^{s}$ is used for testing under the setting of generalized zero-shot scenarios.

Assuming a training set $\mathcal{T} = \left \{ \left ( x_{i},c_{i} | x\subset \mathcal{X}, c\subset \mathcal{C}^{s}      \right )  \right \} $ where $\mathcal{X}$ is the input image space and $c$ belongs to the seen composition label set, CZSL aims to train a model $\mathcal{M}:\mathcal{X} \to \mathcal{C} ^{t}$ to predict compositions in the test samples space $\mathcal{C} ^{t}$. 
If $\mathcal{C} ^{t} \cap \mathcal{C} ^{s} \equiv \phi $, the model only needs to predict unseen compositions.
Following the setting of Generalized ZSL~\cite{xian2018zero}, testing samples contain seen and unseen compositions, i.e., $\mathcal{C} ^{s} \cup \mathcal{C}^{u} $ in this paper.

\subsection{Prompt Initialization}
Since combining multiple senses helps us better understand and analyze new information, human learning is essentially multi-modal, so vision language models (VLMs) are more consistent with human cognition of the world.
CLIP~\cite{AlecRadford2021LearningTV} is a pioneering work of VLMs on visual tasks that promotes the rapid development of computer vision in many areas.
Notably, CLIP has been pretrained with approximately 400 million image-text pairs sourced from the internet.

Our model still follows the architecture of CLIP, whose entire model consists of an image encoder and a text encoder.
Specifically, the image encoder is either a vision transformer or a convolutional neural network, and the text encoder is composed of multiple transformer layers.
The two separate encoders are frozen during the training phase, and they are denoted as $E_{v}$ and $E_{t}$ to represent a vision encoder and a text encoder, respectively.

Prompts exist in CLIP as a photo of \textit{[class]}, where \textit{[class]} is the collection of all classes under the dataset.
In detail, prompts are converted to tokens for every work through passing the tokenizer, and then the embedding function maps them into vocabulary $\mathcal{P}$, which consists of all embeddings of prompts.
Next, the text encoder $E_{t}$ would extract textual representations $f_{t}$ based on them.
Also, image features $f_{v}$ are computed from the image encoder similarly.
The formulation of this process could be described as follows:
\begin{equation}
    f_{v} = Norm(E_{v}(x)), f_{t} = Norm(E_{t}(p)),
\end{equation}
where $x$ represents the input query image in $\mathcal{T}$ and $p$ denotes the prompt in $\mathcal{P}$.
To limit $f_{t}$ and $f_{v}$ to a standard range, they need to be normalized, and $Norm(\cdot)$ is denoted as:
\begin{equation}
    Norm(\cdot) = \frac{f}{||f||}.
\end{equation}
We then take the cosine similarities between the query image representation and the text representation to compute the final prediction, which could be formulated as follows:
\begin{equation}
\label{equ:sim}
    Sim(\frac{y={(a,o)}}{x;\theta } )=\frac{exp(f_{v}\cdot f_{t})}{\sum _{(\bar{a},\bar{o} ) \in \mathcal{C}}exp(f_{v} \cdot  f_{t}) } .
\end{equation}

Since entities in CZSL exist as compositions of state and object, the prompt can be set to "\textit{a photo of} \textit{[state] [object]}" under this framework of CLIP.
Obviously, \textit{[state] [object]} could be treated as \textit{[class]} and its dimension is $\left | \mathcal{A}\cup \mathcal{O} \right | $.
Apart from this, the two vocabulary \textit{[state]} and \textit{[object]} are learnable during training on the seen text-image pairs.
The prompts are also passed through tokenizer and embedding functions, which could be mapped as follows:
\begin{equation}
    p_{a,o} = \left \{ x_{0},x_{1},\dots, x_{pre},x_{a},x_{o}  \right \},
\end{equation}
where $p_{a,o}$ has prefix part $\left \{ x_{0},x_{1},\dots, x_{pre}\right\}$ and vocabulary part $\left \{ x_{a}, x_{o}\right \}$.
To represent learnability in $p_{a,o}$, it could be reformulated as:
\begin{equation}
    \xi (p_{a,o}) = \left \{ x_{0},x_{1},\dots, x_{pre},\theta_{a},\theta_{o}  \right \},
\end{equation}
where $\left \{ \theta_{a}, \theta_{o} \right \}$ is the learnable parameters in all prompts. So we only need to tune $\left | \mathcal{A}\cup \mathcal{O} \right | \times d$ parameters, where $d$ is the dimension of embeddings.
As a result, $f_{t}$ needs to be converted to $Norm(E_{t}(\xi (p_{a,o})))$.
Finally, we can minimize the cross entropy loss for classification:
\begin{equation}
    \mathcal{L}= -\frac{1}{\left | \mathcal{C}^{s} \right | }\sum_{(x,y)\in \mathcal{C}^{s} }log \left ( Sim (\frac{y={(a,o)}}{x;\theta } )\right ).
\end{equation}

\subsection{Disentangled and Recurrent Prompt Tuning}
An intrinsic relationship exists between \textit{[state]} and \textit{[object]}, causing entanglement in the two vocabularies.
For a simple and vivid example, \textit{[object]} has \textit{tiger} and \textit{apple}, \textit{[state]} has \textit{old} and \textit{ripped}, \textit{old} can only be combined with \textit{tiger}, and \textit{apple} can only be \textit{ripped}.
Hence, when the \textit{tiger} embedding is optimized, the \textit{old} embedding is optimized indirectly, regardless of the \textit{ripped}.
As the diagram shown in Fig.~\ref{fig:moti-pull}, such traction force among entanglements largely causes the associated states and objects to accommodate each other during parameter updates, trapping them in a non-optimal space.
Our prompt disentangled tuning framework could inhibit traction and lead these prompts to optimal prompts space.

\textbf{Progressive tuning strategy.} In order to learn more independent and clean feature spaces for states and objects, the entire tuning process is restructured into three significant phases in our model.
It is reasonable to believe that if we freeze the update process of state parameters related to objects, then object embeddings would more independently learn their own optimal parameters, and vice versa.
Specifically, we denote the two learnable vocabularies \underline{\textit{[state]}} and \underline{\textit{[object]}}, also with the frozen vocabularies \textit{[state]} and \textit{[object]}.
As a result, the tuning process could be formulated as triplet status:
\textit{[state]}\underline{\textit{[object]}},
\underline{\textit{[state]}}\textit{[object]},
\underline{\textit{[state][object]}}, and we denote $St$ to represent them as follows:
\begin{equation}
    St=\left \{ o: \textit{[state]}\underline{\textit{[object]}}, a: \underline{\textit{[state]}}\textit{[object]}, ao: \underline{\textit{[state][object]}} \right \},
\end{equation}
where the three states could be abbreviated as $St[o]$, $St[a]$ and $St[ao]$. These three states constitute a status machine, which controls the updating status of model parameters.

Let's analyze the operation process of the status machine in detail: Given an initial state $T_{St}$ and a number of run epochs $K$, the model would run $K$ epochs in this initial state and then transition to the next state, which is selected from the other two states.
The transition of states should not be randomly selected, but maintain a certain regularity, so that the mode could gradually learn well.
In \textbf{\textit{DRPT}}, we set each state to run a fixed $K$ epochs before switching to the next state, and there is independence between states.
Notably, this tuning process disentangles the embeddings of states and objects to some extend, avoiding them being unable to learn better parameters due to entanglement.
By freezing the parameters of some prompt embeddings, the other embeddings could learn their feature space more independently, which is a simple and efficient method.
We set the complete execution of the three states of the model as a round.
When the epoch is small, one round is not enough to update the parameters of the model. Therefore, when one round is completed, the next round will be executed periodically.

\begin{table*}[!htb]\centering
\caption{Comparative analysis of three datasets UT-Zappos~\cite{yu2014fine}, AO-Clevr~\cite{atzmon2020causal} and C-GQA~\cite{naeem2021learning} in our used environment. } 
\label{tab:dataset}
\resizebox{0.85\textwidth}{!}{
\begin{tabular}{ccccccccccccccccc}
\hline
\multirow{2}{*}{\textbf{Dataset}} & \multicolumn{4}{c}{\textbf{Composition}}                     &         &  & \multicolumn{2}{c}{\textbf{Training}} &  & \multicolumn{3}{c}{\textbf{Validation}} &  & \multicolumn{3}{c}{\textbf{Test}}      \\ \cline{2-6} \cline{8-9} \cline{11-13} \cline{15-17} 
                                  & $|\mathcal{A}|$ & $|\mathcal{O}|$ & $Ent_{avg}$ & $Ent^{a}_{var}$ & $Ent^{o}_{var}$ &  & \textit{Cs}        & \textit{i}       &  & \textit{Cs}  & \textit{Cu} & \textit{i} &  & \textit{Cs} & \textit{Cu} & \textit{i} \\ \hline
UT-Zappos                         & 16         & 12         & 0.43            & 9.1              & 8.2     &  & 83                 & 22998            &  & 15           & 15          & 3214       &  & 18          & 18          & 2914       \\
AO-Clevr                          & 8          & 3          & 0.67            & 11.1             & 11.3    &  & 16                 & 102482           &  & 4            & 4           & 38870      &  & 4           & 4           & 38648      \\
C-GQA                             & 413        & 674        & 0.02            & 1092.8           & 111.7   &  & 5592               & 26920            &  & 1252         & 1040        & 7280       &  & 888         & 923         & 5098       \\ \hline
\end{tabular}}
\end{table*}

\textbf{Explanation.} In Fig.~\ref{fig:framework}, we take a demonstration analysis of the prompts in the \textbf{\textit{DRPT}} training process for explanation.
At the first step $St[o]$, the large entanglement state prompt would be updated to a great extend, causing them more closer to related object prompts, and then related object prompts would be updated similarly in $St[a]$.
Finally, these prompts would be joint tuned in $St[ao]$.
Obviously, the prompts of large entanglement would shift more than joint tuning method.
Notably, \textit{[object]} is pre-trained on 400M text-image pairs, which is closer to the real prompts distribution in the combination, that is, the parameters of the combination should be inclined to objects.
The experiment verifies that the overall prompts space would learn better when taking $St[o]$ as the starting state.

\textbf{Entanglement Quantization.} In order to further study the entanglement of state and object, we design two new metrics to quantify the entanglement between them.
The first is the average entanglement rate, $Ent_{avg}$, which is used to evaluate the average entanglement degree between state and object in the current data set. Its formula is defined as follows:
\begin{equation}
    Ent_{avg} = \frac{|\mathcal{C}^{s}|}{|\mathcal{A} | \times |\mathcal{O}|},
\end{equation}
where $|\mathcal{C}^{s}|$ denotes the number of seen compositions in training dataset and $|\mathcal{A} | \times |\mathcal{O}|$ represents the number of all compositions in the whole dataset.
In general, the compositions of test sets are agnostic, so we use the visible compositions of training sets to estimate the average entanglement rate.
Hence, the larger this metric is, the stronger the entanglement between state and object in this dataset is.
Apart from the metric for measuring entanglement rate, we also introduce two variances for entanglement: $Ent_{var}^{a}$ and $Ent_{var}^{o}$.
In a dataset, an object may be associated with 100 states, while some objects are combined with only one state, so the entanglement rate is not evenly distributed among each object or state.
And $Ent_{var}$ could be formulated as follows:
\begin{equation}
\label{equ:entao}
    Ent^{a}_{var}=\sum_{a\subset \mathcal{A}} \frac{ (Ent^{a}-Ent_{avg})^2}{|\mathcal{A}|}, Ent^{o}_{var}=\sum_{o\subset \mathcal{O}} \frac{ (Ent^{o}-Ent_{avg})^2}{|\mathcal{O}|},
\end{equation}
where $Ent^{a}$ and $Ent^{o}$ denote the entanglement of each state and object.
For example, if the tiger has two states describing it, then the tiger's entanglement is 2.
The larger the variance $Ent_{var}^{a}$ and $Ent_{var}^{o}$, the more uneven the distribution of entanglement in the data set, and the larger the long-tail effect.

To effectively mitigate the problem of uneven optimization caused by long-tail distribution, we reweight the cross entropy loss of classification loss. The weight of each composition could be computed by:
\begin{equation}
\label{euq:w-}
    w_{-} = 1 + \alpha \times (1 - Norm(Ent^{a} \times Ent^{o})),
\end{equation}
\begin{equation}
\label{euq:w+}
    w_{+} = 1 + \alpha \times Norm(Ent^{a} \times Ent^{o}),
\end{equation}
which represents the entanglement weight between state $a$ and object $o$.
If $Ent_{avg}$ is low, we need $w_{+}$ to enhance the parameters updating of strong entanglements and $w_{-}$ is designed to suppress the traction force of strong entanglement prompts.
% It is necessary to give larger loss weight to the composition with high entanglement, so that the model pays more attention to the learning of this part. Meanwhile, we also test the results of the experiment by giving smaller weights.
Finally, the composition classification cross entropy loss could be formulated as follows:
\begin{equation}
\label{equ:com}
    \mathcal{L}= -\frac{w}{\left | \mathcal{C}^{s} \right | }\sum_{(x,y)\in \mathcal{C}^{s} }log \left ( Sim (\frac{y={(a,o)}}{x;\theta } )\right ).
\end{equation}

To show the flow of the algorithm more intuitively, the training procedure of \textbf{\textit{DRPT}} is shown in Algorithm.~\ref{alg:training}.

% $St[o]$ and $St[a]$ will enable the state and object embedding to learn more independent feature parameters, but joint training of $St[so]$ may cause them to be entangled again. 
% To tackle with this challenge, we design regularization loss terms to suppress the entangled joint update in $St[ao]$ stage.
% And the regularization loss could be computed as:
% \begin{equation}
% \label{equ:reg}
%     \mathcal{L}_{reg}=\sum_{a\in\mathcal{A} } Ent^{a}\times|| \theta _{a} - \theta _{a}^{St[a]}||+\sum_{o\in\mathcal{O} } Ent^{o}\times|| \theta _{o} - \theta _{o}^{St[o]}||.
% \end{equation}
% Finally, the overall loss could be described as follows:
% \begin{equation}
% \label{equ:L}
%     \mathcal{L} = \mathcal{L}_{com} + \beta \times \mathcal{L}_{reg},
% \end{equation}
% where $\beta$ is a hyper-parameter that controls the loss of the regular term and is set to 0.1 by default.
% To show the flow of the algorithm more intuitively, the training procedure of \textbf{\textit{DRPT}} is shown in Algorithm.~\ref{alg:training}.

\begin{algorithm}[t]  %其中这里面不能有H不然会报错，不过不影响结果
 \caption{Training procedure of \textbf{\textit{DRPT}} ($o \to a \to ao$)}%算法名字
 \label{alg:training}
 \LinesNumbered %要求显示行号
 \KwIn{Training set $\mathcal{T} = \left \{ \left ( x_{i},c_{i} | x\subset \mathcal{X}, c\subset \mathcal{C}^{s}      \right )  \right \} $, Model $\mathcal{M}$}%输入参数
 \textbf{Initialize:} Object set $\mathcal{O}$, State set $\mathcal{A}$, 
    \\Training status $T_{St}=St[o]$, Round range $K$,
    \\Prompts set $P = \{\xi (p_{a,o})|a\in\mathcal{A}, o\in\mathcal{O}\}$ %\;用于换行
 \\
 \KwOut{Optimal prompts embeddings;}
 Compute $Ent^{a}_{var}$ and $Ent^{o}_{var}$ via Equ.~\ref{equ:entao}\;
 Freeze $\theta_{a}$, train $\theta_{o}$\;
\While{not converged}{
    Sample a batch from $\mathcal{T}$ \;
    \For{k = 1 to $K$}{
    \For{samples within the batch}{
        $f_{v} = Norm(E_{v}(x))$, $f_{t} = Norm(E_{t}(p))$ \;
        Compute similarity via Equ.~\ref{equ:sim} \;
        Compute weight $w$ via Equ.~\ref{euq:w-} or~\ref{euq:w+} \; 
        Reweight the loss $\mathcal{L}$ via Equ.~\ref{equ:com} \;
        Update parameters in $\{\xi (p_{a,o})|a\in\mathcal{A}, o\in\mathcal{O}\}$ \;
        }
    }
    // Update tuning Status $T_{St}$\;
    \Switch{$T_{St}$}{
        \Case{$St[o]$}{$T_{St}=St[a]$, freeze $\theta_{o}$, train $\theta_{a}$\;}
        \Case{$St[a]$}{$T_{St}=St[ao]$, train $\theta_{a}$ with $\theta_{o}$\;}
        \Case{$St[ao]$}{$T_{St}=St[o]$, freeze $\theta_{a}$, train $\theta_{o}$\;}
    }
}
\end{algorithm}

\subsection{Inference}
For CZSL, there are vast unseen compositions during inference, so we recompose the fine-tuned attribute and object vocabulary in the prompt.
We have learned the state and object prompt parameters in the training phase, and the test prompts are rearrangement of them.
The most likely compositions of state and object could be calculated by:
\begin{equation}
    \hat{y} = argmax (Sim(\frac{y={(a,o)}}{x:\theta } )), y \in \mathcal{C}^s \cup \mathcal{C}^u.
\end{equation}

\section{Experiment}
In this section, all required datasets and evaluation protocols would be described concretely. Also, we present extensive comparable experiments with vast state-of-the-art algorithms and the implementation details. Meanwhile, ablation studies demonstrate the high effectiveness of our proposed method.

\subsection{Experimental Setup}

Our experimental setup follows the settings of other state-of-the-art methods of CZSL and ensures the fairness of the experiment, which includes the dataset and the evaluation metrics.

\begin{table*}[!htb]\centering
\caption{Results of \textbf{\textit{DRPT}} on UT-Zappos, AO-Clevr and C-GQA. \textit{Seen} and \textit{Unseen} are the predicted accuracy evaluated on seen and unseen compositions. \textit{HM} is the harmonic mean of \textit{Unseen} and \textit{Seen} and \textit{AUC} is the area under the curve. The best results are in bold and the second results are in underlined.} 
\label{tab:closed-world}
\resizebox{0.95\textwidth}{!}{
\begin{tabular}{ccccccccccccccc}
\hline
\multirow{2}{*}{\textbf{Method}} & \multicolumn{4}{c}{\textbf{UT-Zappos}}                               &  & \multicolumn{4}{c}{\textbf{AO-Clevr}}                                 &  & \multicolumn{4}{c}{\textbf{C-GQA}}                                     \\ \cline{2-5} \cline{7-10} \cline{12-15} 
                        & Seen          & Unseen        & HM            & AUC         &  & Seen         & Unseen        & HM            & AUC           &  & Seen          & Unseen        & HM            & AUC          \\ \hline
AoP~\cite{nagarajan2018attributes}                     & 59.8          & 54.2          & 40.8          & 25.9        &  & 95.5         & 85.5          & 64.8          & 65.8          &  & 17.0          & 5.6           & 5.9           & 0.7          \\
LE+~\cite{naeem2021learning}                     & 53.0          & 61.9          & 41.0          & 25.7        &  & 95.7         & 99.2          & 92.3          & 93.5          &  & 18.1          & 5.6           & 6.1           & 0.8          \\
TMN~\cite{purushwalkam2019task}                     & 58.7          & 60.0          & 45.0          & 29.3        &  & 96.1         & 93.9          & 45.0          & 29.3          &  & 23.1          & 6.5           & 7.5           & 1.1          \\
SymNet~\cite{YongLuLi2020SymmetryAG}                  & 49.8          & 57.4          & 40.4          & 23.4        &  & 87.1         & 97.8          & 71.8          & 74.2          &  & \textbf{30.9}          & 13.3          & 13.5          & 3.1          \\
CompCos~\cite{mancini2021open}                 & 59.8          & 62.8          & 43.1          & 28.7        &  & 96.3         & 99.1          & 94.5          & 94.2          &  & \underline{30.7}         & 12.2          & 12.8          & 2.9          \\
IVR~\cite{zhang2022learning}                     & 56.9          & 65.5          & 46.2          & 30.6        &  & \underline{97.1}         & 99.3          & \underline{95.1}          & \underline{95.6}          &  & 27.3             & 10.0             & 10.9             & 2.2            \\
SCEN~\cite{li2022siamese}                     & 63.5          & 63.1          & \underline{47.8}          & 32.0        &  & -            & -             & -             & -             &  & 28.9          & 25.4          & 17.5          & 5.5          \\
CLIP~\cite{AlecRadford2021LearningTV}                    & 15.8          & 49.1          & 15.6          & 5.0         &  & 61.4         & 99.7          & 60.3          & 51.8          &  & 7.5           & 25.0          & 8.6           & 1.4          \\
CoOp~\cite{KaiyangZhou2021LearningTP}                    & 52.1          & 49.3          & 34.6          & 18.8        &  & 95.8         & \textbf{99.8}          & 94.6          & 95.2          &  & 20.5          & 26.8          & 17.1          & 4.4          \\
CSP~\cite{nayak2022learning}                     & \underline{64.2}          & \underline{66.2}          & 46.6          & \underline{33.0}        &  & 96.6         & 99.5          & \textbf{95.7} & 95.3          &  & 27.4 & \underline{27.1}          & \underline{19.9} & \underline{5.9}          \\ \hline
\textbf{\textit{DRPT}}                    & \textbf{64.5} & \textbf{69.4} & \textbf{52.3} & \textbf{38.5} &  & \textbf{100} & \textbf{99.8} & 94.5          & \textbf{97.3} &  & 29.2 & \textbf{28.7} & \textbf{20.5} & \textbf{6.5} \\ \hline
\end{tabular}
}
\end{table*}

\textbf{Datasets.} 
For CZSL, the three more commonly used data sets are MIT-States~\cite{isola2015discovering}, UT-Zappos~\cite{yu2014fine} and C-GQA~\cite{naeem2021learning}, where the average entanglement rates of MIT-States and C-GQA are close.
Therefore, we replace MIT-States with the AO-Clevr~\cite{atzmon2020causal} dataset with a higher average entanglement rate to better analyze the relationship between entanglement and tuning.
Another significant reason for abandoning Mit-States is that it's labeled automatically, containing much noise.
Also, some expressions of attributes or labels are confused, as described in \cite{zhang2022learning}.

As a result, our experiments are conducted on three challenging real-world benchmark datasets: UT-Zappos~\cite{yu2014fine}, AO-Clevr~\cite{atzmon2020causal} and C-GQA~\cite{naeem2021learning} respectively.
Specifically, AO-Clevr contains 180K natural images with 8 states and 3 objects
The shoes dataset UT-Zappos consists of 29126 images, composed of 16 states and 12 objects.
Also, C-GQA, the most-paired dataset for CZSL, contains 453 states and 870 objects, totaling 39298 images.
As for compositions, AO-Clevr includes 16 seen compositions, 4 unseen for validation, and 4 unseen for testing, following the split settings in \cite{atzmon2020causal} .
UT-Zappos consists of 83 seen and 15/18 (validation/test) unseen compositions.
For C-GQA, it is constrained to 5592 compositions for training, 1040 compositions for validation, and 923 for testing.
We calculate the entanglement average rate and variance, then construct this information into a dataset statistics table, as shown in Tab.~\ref{tab:dataset}.

\textbf{Metrics.}
In our experiments, we calculate the prediction accuracy based on the compositions.
Same as the settings of previous work~\cite{mancini2021open}, we compare the following metrics with other state-of-the-art methods:
\textit{Seen}, \textit{Unseen}, \textit{HM}, \textit{AUC}.
To be specific, \textit{Seen (S)} and \textit{Unseen (U)} denote the accuracy tested only on seen compositions and unseen compositions respectively.
Also, we can calculate \textit{Harmonic Mean (HM)} of \textit{S} and \textit{U} metrics.
Since zero-shot models have inherent bias for seen compositions, we can draw a seen-unseen accuracy curve at different operating points with the bias from $-\infty $ to $+\infty$ to compute the \textit{Area Under the Curve (AUC)}.

\textbf{Implementation Details.}\
We implement \textbf{\textit{DRPT}} with PyToch 1.12.1~\cite{paszke2019pytorch} and optimize it by Adam optimizer over the forementioned three challenging datasets. The image encoder and text encoder are both based on the pretrained CLIP Vit-L/14 model, and the entire model are trained and evaluated on 1 $\times$ NVIDIA RTX 3090 GPU.
Also, the batch size of our model is 128 during training on all datasets. More parameter details could be seen in Appendix.~\ref{appendix:hyper}.

\subsection{Main Results}
To verify the effectiveness of \textbf{\textit{DRPT}}, we compare our method with extensive prior compositional zero-shot learning algorithms, including vision based methods AoP~\cite{nagarajan2018attributes}, LE+~\cite{naeem2021learning}, TMN~\cite{purushwalkam2019task}, SymNet~\cite{YongLuLi2020SymmetryAG}, CompCos~\cite{mancini2021open}, IVR~\cite{zhang2022learning} and SCEN~\cite{li2022siamese}, and vision-language based models CLIP~\cite{AlecRadford2021LearningTV}, CoOp~\cite{KaiyangZhou2021LearningTP}, CSP~\cite{nayak2022learning}.
We analyze the four classical metrics \textit{Seen, Unseen, HM} and \textit{AUC} on UT-Zappos~\cite{yu2014fine}, AO-Clevr~\cite{atzmon2020causal} and C-GQA~\cite{naeem2021learning}, and the experiment results are shown in Tab. \ref{tab:closed-world}. The best results are in \textbf{bold} and the second results are in \underline{underlined}.

In Tab.~\ref{tab:closed-world}, \textbf{\textit{DRPT}} outperforms almost all reported results.
Specifically, \textbf{\textit{DRPT}} improves \textit{AUC} by +5.5\%, +1.7\% and +0.6\% on UT-Zappos, AO-Clevr and C-GQA respectively, and improves \textit{HM} by +4.5\% on UT-Zappos.
Meanwhile, \textbf{\textit{DRPT}} improves CSP by +0.3\% on UT-Zappos for \textit{Seen} and +1.8\% on AO-Clevr for \textit{Seen}.
On overall metrics, \textbf{\textit{DRPT}} only decreases on AO-Clevr for \textit{HM} and C-GQA for \textit{Seen}.
For the \textit{Unseen} accuracy, \textbf{\textit{DRPT}} improves +3.2\% and +1.6\% on UT-Zappos and C-GQA.
\textbf{\textit{DRPT}} surpasses extensive experiments by large margin, demonstrating our method can effectively alleviate the problem that the gradient update of the model is misled due to the traction force of entanglements.

Apart from this, we report the unseen-seen accuracy curve on UT-Zappos and C-GQA.
As the calibration value increases, the accuracy of classifying unseen pairs rises while that of seen pairs declines.
The \textit{AUC} is the calculated area between the curve and the coordinate axis, and is an important measure of the robustness of the accuracy seen and unseen.
Compared with other representative algorithms, our method keeps a better balance between seen and unseen pairs on both datasets.
The \textcolor{blue}{blue} line denotes our method and each scatter data is greater than other algorithms on UT-Zappos and surpasses previous state-of-the-art method CSP greatly.
All of these experimental results sufficiently demonstrate the superiority of our proposed method to a great extend.

\begin{figure*}[t]
  \centering
  \includegraphics[scale=0.45]{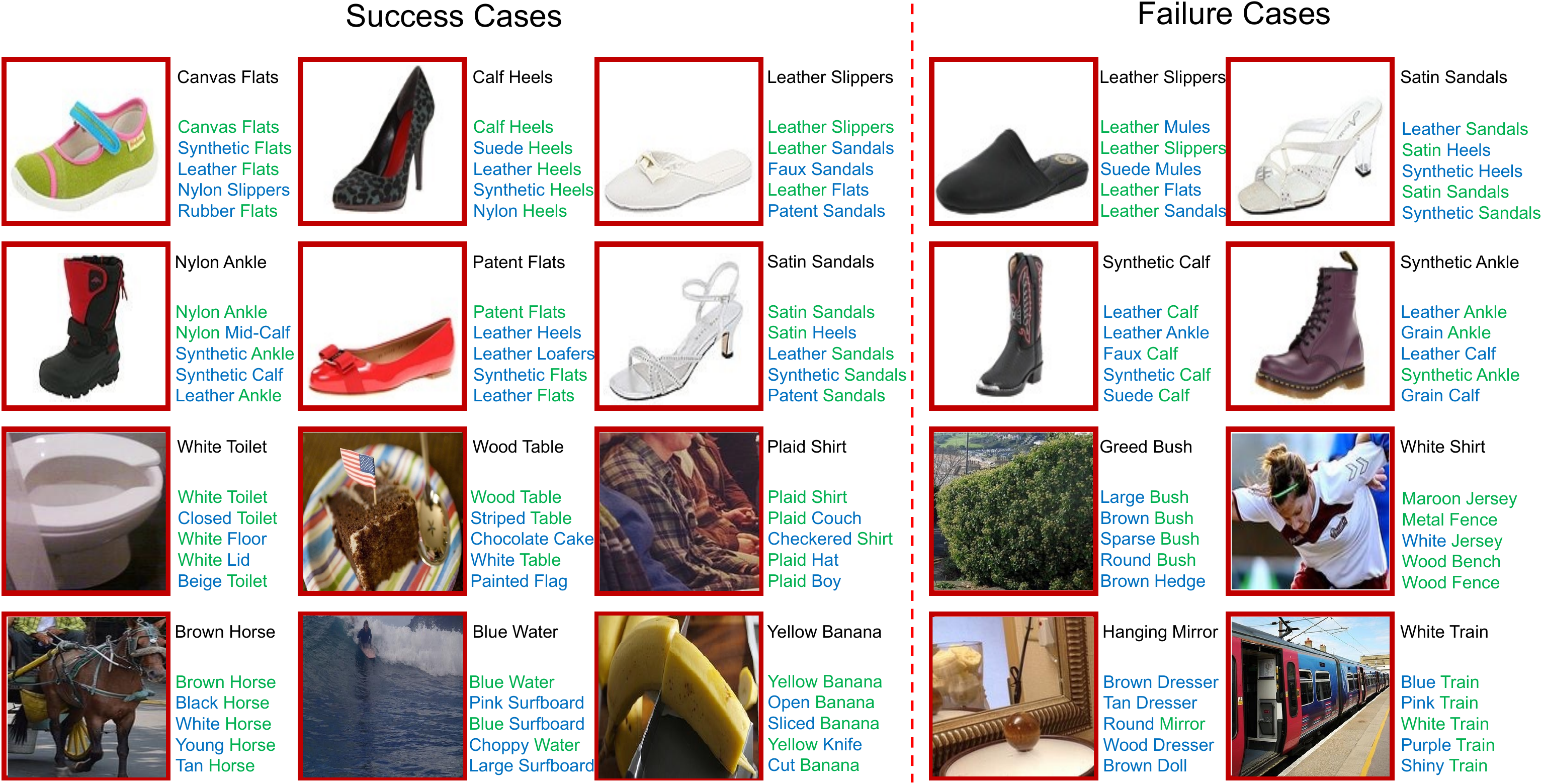}
   \caption{Qualitative results for Image-to-Text retrieval. We evaluate top-5 predictions for some cases on UT-Zappos and C-GQA. For the failure cases, \textcolor{blue}{blue} denotes the wrong prediction and all images are randomly selected.}
   \label{fig:demo_img}
\end{figure*}

\begin{figure}[t]
  \centering
  \includegraphics[scale=0.5]{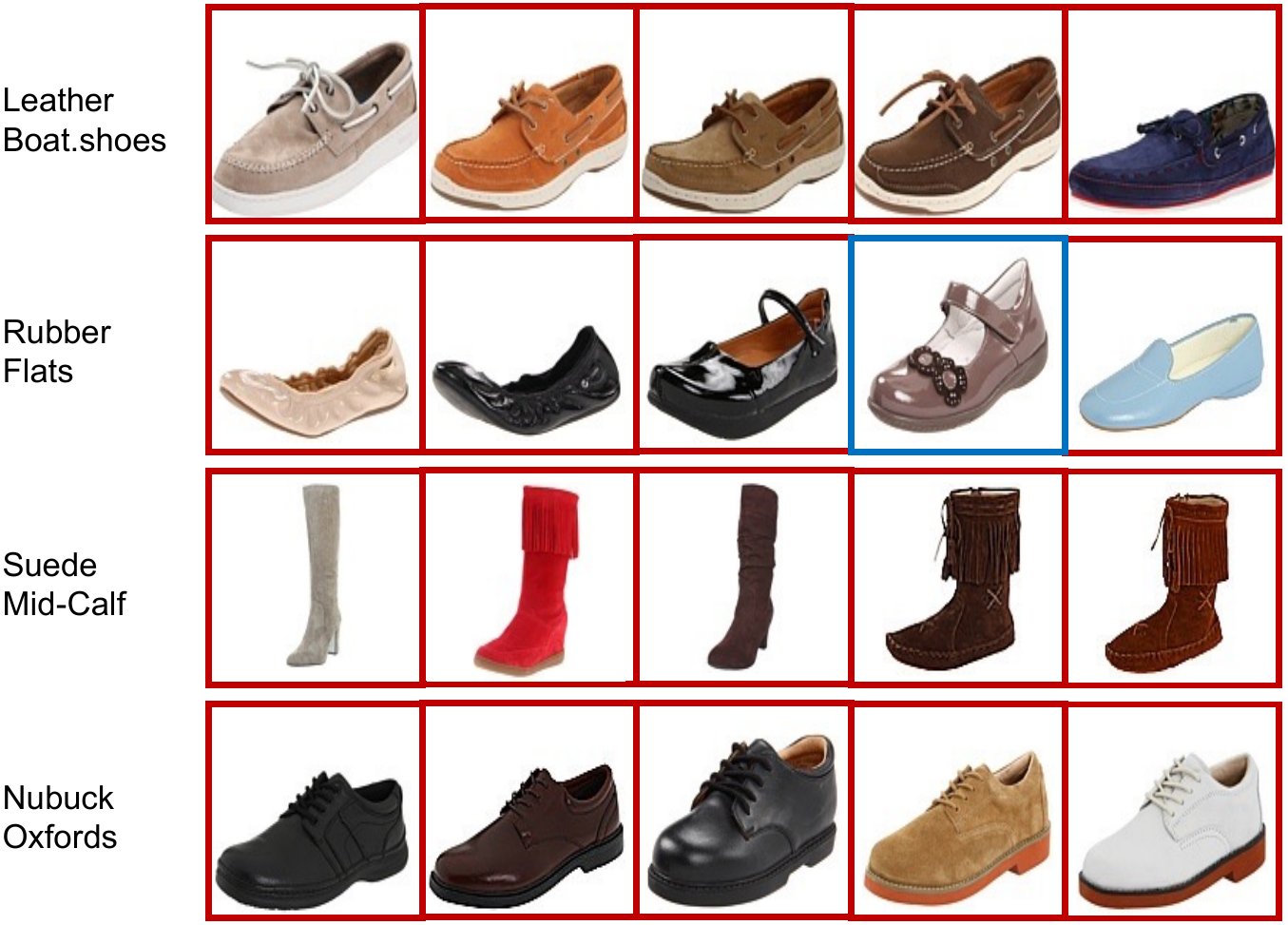}
   \caption{Qualitative results for Text-to-Image retrieval. We evaluate the top-5 predictions for some cases on UT-Zappos. For the failure cases, \textcolor{blue}{blue} denotes the wrong prediction and all texts are randomly selected.}
   \label{fig:demo_txt}
\end{figure}

\begin{figure}
  \centering
  \subfigure[\textit{AUC} curve on UT-Zappos]{
  \begin{minipage}{0.23\textwidth}
    \centering
    \includegraphics[scale=0.3]{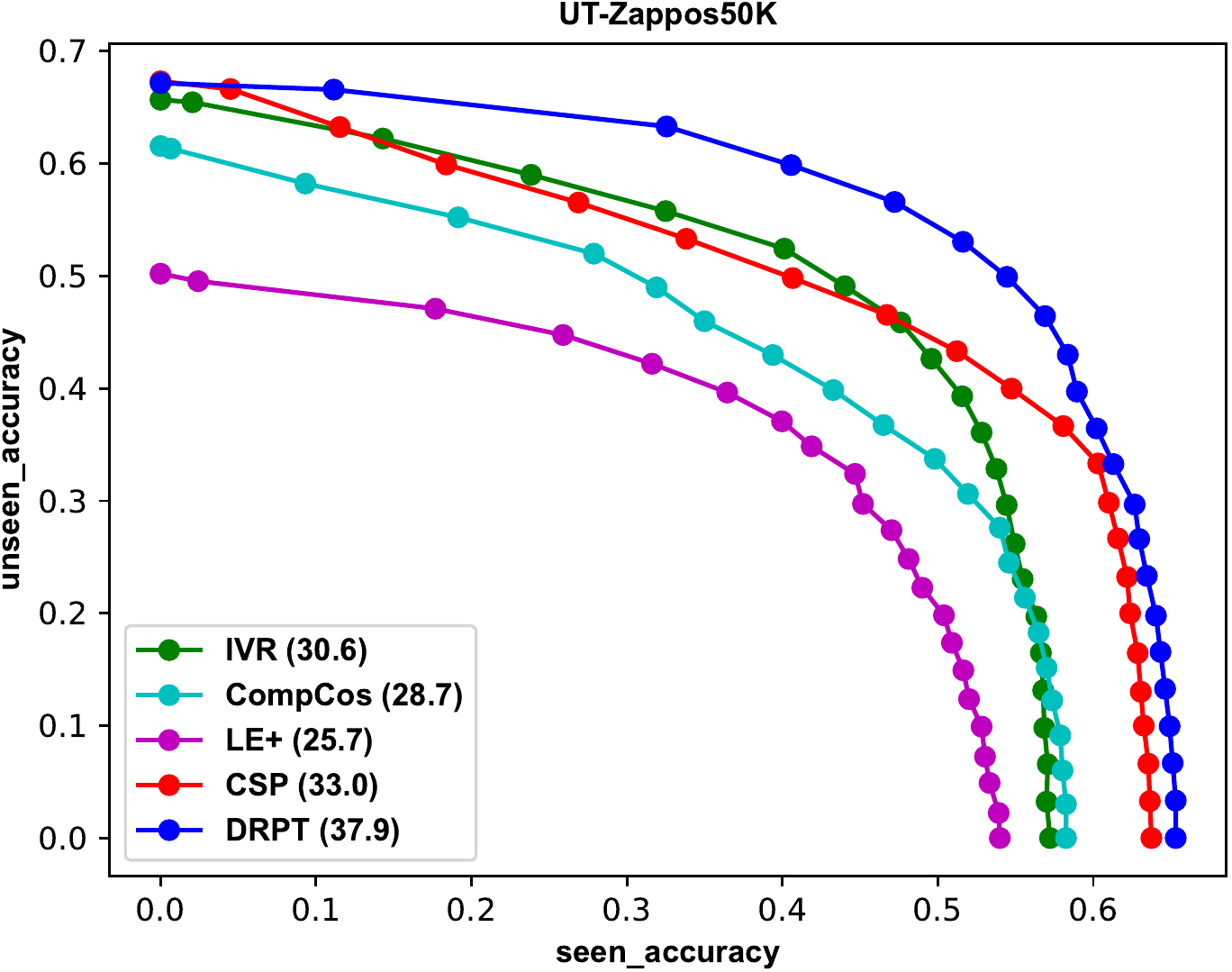}
    % \caption{Caption for image1}
    \label{fig:sub1}
  \end{minipage}\hfill}
  \subfigure[\textit{AUC} curve on C-GQA]{
  \begin{minipage}{0.23\textwidth}
    \centering
    \includegraphics[scale=0.3]{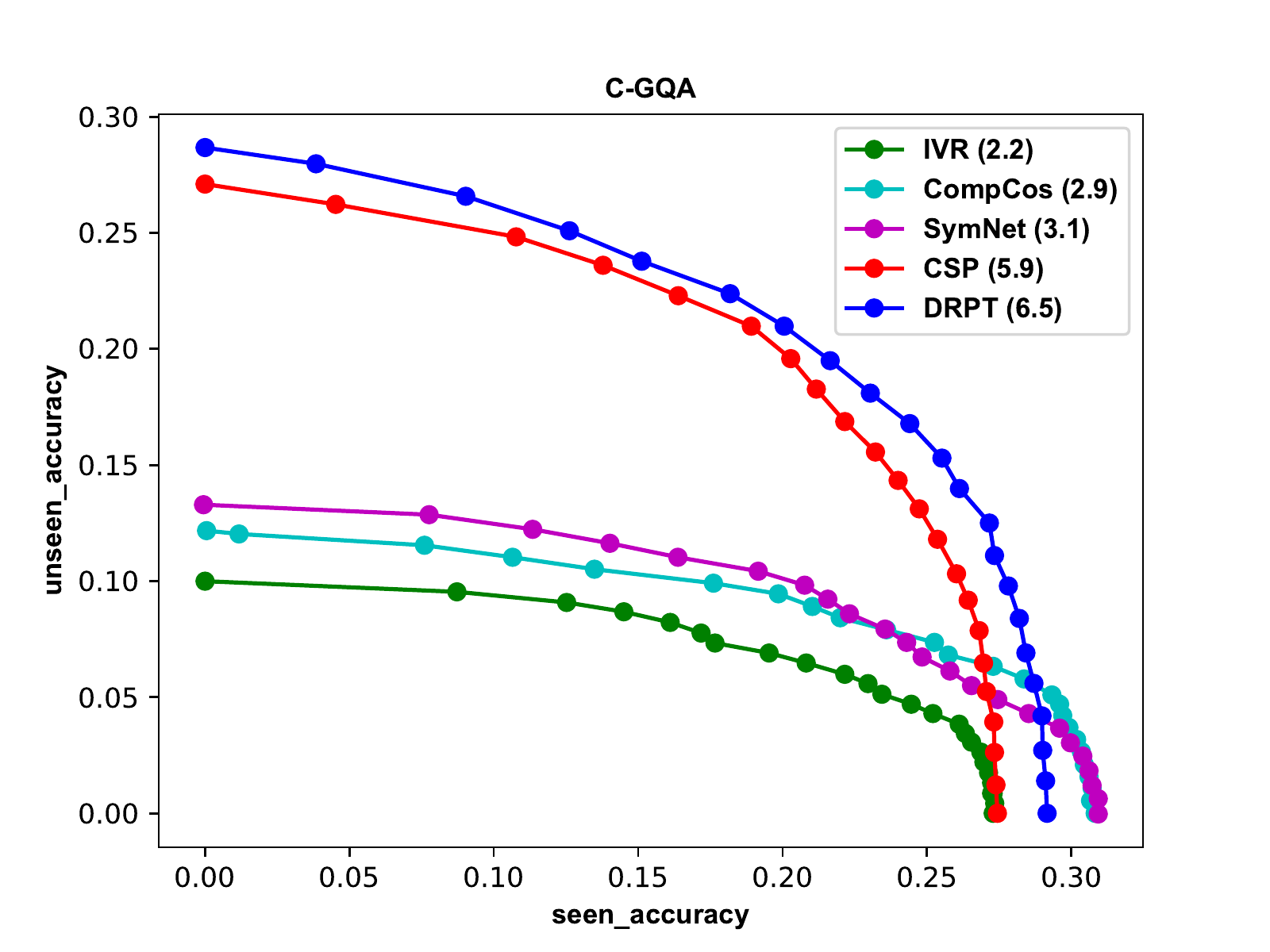}
    % \caption{Caption for image2}
    \label{fig:sub2}
  \end{minipage}}
  \caption{Unseen-seen accuracy curve on two challenging datasets UT-Zappos~\cite{yu2014fine} and C-GQA~\cite{naeem2021learning}, where \textbf{\textit{DRPT}} is in \textcolor{blue}{blue}.
  We compare \textbf{\textit{DRPT}} with other representative algorithms, and \textit{AUC} surpasses these models by large margins, verifying the superiority of our method.}
  \label{fig:AUC}
\end{figure}

\subsection{Ablation Study}
To empirically show the effectiveness of \textbf{\textit{DRPT}}, we conduct extensive experiments and report the results for the ablation study.

\textbf{Ablation on different components on \textbf{\textit{DRPT}}.}
We first evaluate the effectiveness of composition classification weight $w$ for various entanglements and progressive tuning strategy (D\&R-PT) on UT-Zappos and C-GQA.
The results are shown in Tab.~\ref{tab:com}, in which entanglement reweight $w$ and D\&R-PT are tested to evaluate the effectiveness.
Specifically, - denotes the $w_{-}$ for inhibition of entanglements, while + represents the $w_{+}$ for enhancement of entanglements.
It is obvious that D\&R-PT and $w$ could really bring positive effects like +2.4\% and +3.1\% for AUC on UT-Zappos respectively.
As the Tab.~\ref{tab:dataset} shown, UT-Zappos exists as high $Ent_{avg}=0.43$ and C-GQA contains low $Ent_{avg}=0.02$.
To rebalance the effects of entanglements, UT-Zappos needs to be inhibited and C-GQA should be enhanced, and the results verify this assumption.
Given $w_{+}$ for UT-Zappos, the results would be decreased a lot, also with $w_{-}$ for C-GQA.

\textbf{Ablation on different status transition sequences.}
The default initial status in the experiment is $St[o]$, that is, $T_{St}=St[o]$.
Specifically, $St$ has three statuses in total, and there may be 6 transition sequences in a round of statues transition, as shown in Tab.~\ref{tab:seq}.
In these cases, "$a \to o \to ao$" represents the initial tuning status is $St[a]$, followed by $St[o]$ and then $St[ao]$, and and tuning is carried out periodically.
As the results shown in Tab.~\ref{tab:seq}, "$o \to a \to ao$" presents the best results among this status transition sequences and "$o \to ao \to o$" shows the second results.
When $St[ao]$ is advanced, the overall effect is worse because both state and object have not been learned independently.
On the whole, there is a sequence that makes the effect higher than joint tuning.
Moreover, the dynamic transformation of sequences can be further studied in future work, such as dynamic $K$ value and automatic status switching.

\begin{table}[t]
\caption{Ablation study for different components in \textbf{\textit{DRPT} on UT-Zappos and C-GQA. The best results are in \textbf{bold}.}}
\label{tab:com}
\resizebox{0.47\textwidth}{!}{
\begin{tabular}{ccccccccccc}
\hline
\multirow{2}{*}{\begin{tabular}[c]{@{}c@{}}D\&R-PT\end{tabular}} & \multirow{2}{*}{$w$}          & \multicolumn{4}{c}{\textbf{UT-Zappos}} &  & \multicolumn{4}{c}{\textbf{C-QGA}} \\
                                                                                &                             & S       & U       & HM      & AUC      &  & S      & U      & HM     & AUC     \\ \hline
\XSolidBrush                                                     & \XSolidBrush & 64.2        &  66.2       & 46.4        & 33.0         &  & 27.4       & 27.1       & 19.9       & 5.9        \\
\XSolidBrush                                                     & -                           &  58.8       & 62.9        & 43.2        & 28.5         &  & 27.8        & 27.1       & 19.6       & 5.9        \\
\XSolidBrush                                                     & +                           & \textbf{64.9}        & 66.8        &  45.6       & 32.6         &  & 27.9       & 27.1       & 19.5       & 5.9        \\
\Checkmark                                                       & \XSolidBrush &  64.7       & 67.4        & 48.9        & 35.4         &  & 28.3       &  26.9      & 19.7       & 5.9        \\
\Checkmark                                                       & -                           &  64.5       & \textbf{69.4}        & \textbf{52.3}        & \textbf{38.5}         &  & 28.0       &  27.1      & 19.5       &  6.0       \\
\Checkmark                                                       & +                           &  62.7       & 63.6        & 46.8        & 32.6          &  &  \textbf{29.2}       & \textbf{28.7}        & \textbf{20.5}        & \textbf{6.5}         \\ \hline
\end{tabular}
}
\end{table}

\begin{table}[h]
\caption{Ablation study for different training status transition sequences in \textbf{\textit{DRPT} on UT-Zappos and C-GQA. The best results are in \textbf{bold}.}}
\label{tab:seq}
\resizebox{0.47\textwidth}{!}{
\begin{tabular}{cccccccccc}
\hline
\multirow{2}{*}{\textbf{\begin{tabular}[c]{@{}c@{}}Status\\ Sequence\end{tabular}}} & \multicolumn{4}{c}{\textbf{UT-Zappos}} &  & \multicolumn{4}{c}{\textbf{C-GQA}} \\
                                                                                    & S    & U    & HM      & AUC    &  & S    & U    & HM   & AUC   \\ \hline
$a \to o \to ao$                                                                            & 64.7    & 68.4      & 49.0    & 35.0   &  & 28.5         & 25.6          & 19.6     & 5.8      \\
$a \to ao \to o$                                                                             & 63.4    & 70.0      & 46.4    & 33.5   &  & 26.8        & 28.1          & 20     & 6.1      \\
$o \to a \to ao$                                                                            & \textbf{64.5}    & \textbf{69.4}      & \textbf{52.3}    & \textbf{38.5}   &  & \textbf{29.2}        & \textbf{28.7}          & \textbf{20.5}     & \textbf{6.5}      \\
$o \to ao \to a$                                                                            & 65.1    & 67.3      & 50.4    & 36.7   &  & 28.4        & 27.7          & 20.3     & 6.2      \\
$ao \to a \to o$                                                                            & 64.3    & 69.4      & 50.0    & 36.2   &  & 28.2        & 26.2          & 19.7     &   5.9    \\
$ao \to o \to a$                                                                            & 64.2    & 67.9      & 46.5    & 33.7   &  & 28.3        & 27.5          & 19.8     & 6.0      \\ \hline
\end{tabular}}
\end{table}

\subsection{Qualitative Results}
\textbf{Image-to-Text and Text-to-Image Retrieval.} We visualize some qualitative results for seen and unseen compositions with top-5 Image-to-Text retrieval predictions in Fig.~\ref{fig:demo_img}, where the samples are randomly selected from UT-Zappos and C-GQA.
The successful primitive concepts are highlighted in \textcolor{green}{green} and \textcolor{blue}{blue} denotes the wrong prediction.
In successful prediction cases, top-2 to top-5 results can basically guarantee that one state or object can be predicted.
For example, the results of "\textit{Brown Horse}", top-5 results can completely predict "\textit{horse}", just the difference in color evaluation.
In failure cases, some baffling compositions are challenging to be predicted, like "\textit{Greed Bush}" and "\textit{Hanging Mirror}".
In all cases, object is always more predictable than state, which is also consistent with the real-world situation.
We then consider Text-to-Image retrieval predictions, and the results of \textbf{\textit{DRPT}} on UT-Zappos are shown in Fig.~\ref{fig:demo_txt}.
Given a context pair "\textit{Rubber Flats}", the top-5 retrieval images would be generated through \textbf{\textit{DRPT}} and the results also highlighted in \textcolor{green}{green} for success cases and \textcolor{blue}{blue} for failure cases.
Only one failure case in "\textit{Rubber Flats}", existing in top-4 retrieval image, and it's very close to other correct cases, so the model misjudged.
Overall, the retrieval procedure employing our algorithm can produce positive outcomes.

\section{Conclusion}
In this paper, we explore the impact of entanglement between state and object primitives from the perspective of prompt tuning, and propose a novel VLMs based framework termed \textbf{D}isentangled and \textbf{R}ecurrent \textbf{P}rompt \textbf{T}uning (\textbf{\textit{DRPT}}) for compositional zero-shot learning.
Specifically, we first analyze the existence of entanglement between state and object would mislead the gradient update of prompts in VLMs and restrict them from going beyond that local optimum. 
To better improve VLMs for CZSL, we design a progressive tuning strategy to tackle the problems of entanglement.
\textbf{\textit{DRPT}} provides certain guidance for their gradient update by freezing some prompts periodically, and makes the model learn better parameters under the original simple structure.
Additionally, we propose the concepts of average entanglement rate and entanglement variance, and further analyze and verify the effect of entanglement on the model.
Extensive experiments verify the superiority and effectiveness of \textbf{\textit{DRPT}}, towards the goal of CZSL.
Moreover, \textbf{\textit{DRPT}} is an illuminating strategy for both quantifying entanglement and fine-tuning, and we hope more research could be expanded based on it.

\newpage

% \begin{acks}
% \end{acks}

\bibliographystyle{ACM-Reference-Format}
\bibliography{reference}

\end{document}